\documentclass[11pt,a4paper]{article}
\usepackage[hyperref]{naacl2021}
\usepackage{amsmath}
\usepackage{amsfonts}
\usepackage{times}
\usepackage{latexsym}
\usepackage{xspace}
\usepackage{enumitem}
\usepackage{comment}
\usepackage{url}
\usepackage{booktabs}
\usepackage{graphicx}
\usepackage{multirow}
\usepackage{cleveref}
\crefformat{section}{\S#2#1#3}
\crefformat{subsection}{\S#2#1#3}
\crefformat{subsubsection}{\S#2#1#3}
\crefrangeformat{section}{\S\S#3#1#4 to~#5#2#6}
\crefmultiformat{section}{\S\S#2#1#3}{ and~#2#1#3}{, #2#1#3}{ and~#2#1#3}
\usepackage{refstyle}
\Crefformat{figure}{#2Fig.~#1#3}
\Crefmultiformat{figure}{Figs.~#2#1#3}{ and~#2#1#3}{, #2#1#3}{ and~#2#1#3}
\Crefformat{table}{#2Tab.~#1#3}
\Crefmultiformat{table}{Tabs.~#2#1#3}{ and~#2#1#3}{, #2#1#3}{ and~#2#1#3}
\Crefformat{appendix}{\S#2#1#3}

\newcommand{\stitle}[1]{\vspace{2ex} \noindent{\bf #1}}
\newcommand{\modelname}[0]{\texttt{JEANS}\xspace}

\def\lang{\mathcal{L}}
\def\bhline{\specialrule{.2em}{0em}{0em}}

\def\hitsone{\mathit{H}\mbox{@}1}
\def\hitsfive{\mathit{H}\mbox{@}5}
\def\hitsten{\mathit{H}\mbox{@}10}
\def\hitsk{\mathit{H}\mbox{@}p}

\def\mrr{\mathit{MRR}}

\usepackage{cleveref}
\crefformat{section}{\S#2#1#3}
\crefformat{subsection}{\S#2#1#3}
\crefformat{subsubsection}{\S#2#1#3}
\crefrangeformat{section}{\S\S#3#1#4 to~#5#2#6}
\crefmultiformat{section}{\S\S#2#1#3}{ and~#2#1#3}{, #2#1#3}{ and~#2#1#3}
\Crefformat{table}{#2{}{Table} #1{}#3}
\Crefformat{figure}{#2{}{Figure} #1{}#3}

\title{Cross-lingual Entity Alignment with Incidental Supervision}

\author{Muhao Chen$^{1,2}$\thanks{\indent Indicating equal contributions.} , Weijia Shi$^{3*}$, Ben Zhou$^1$, Dan Roth$^1$\\
  $^1$Department of Computer and Information Science, UPenn\\
  $^2$Viterbi School of Engineering, USC\\
  $^3$Department of Computer Science, UCLA\\
  \texttt{\{muhao,xyzhou,danroth\}@seas.upenn.edu};  \texttt{swj0419@cs.ucla.edu}\\}

\date{}

\begin{document}
\maketitle
\begin{abstract}
Much research effort has been put to multilingual knowledge graph (KG) embedding methods to address the entity alignment task, which seeks to match entities in different language-specific KGs that refer to the same real-world object.
Such methods are often hindered by the insufficiency of seed alignment provided between KGs. 
Therefore, we propose an incidentally supervised model, \modelname 
\includegraphics[height=1em]{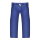}, 
which jointly represents multilingual KGs and text corpora in a shared embedding scheme,
and seeks to improve entity alignment with incidental supervision signals from text.
\modelname first deploys an entity grounding process to combine each KG with the monolingual text corpus.
Then, two learning processes are conducted: (i) an \emph{embedding learning process} to encode the KG and text of each language in one embedding space, and (ii) a self-learning based \emph{alignment learning process} to iteratively induce the matching of entities and that of lexemes between embeddings.
Experiments on benchmark datasets show that \modelname leads to promising improvement on entity alignment with incidental supervision, and significantly outperforms state-of-the-art methods that solely rely on internal information of KGs.\footnote{Software and resources are available at \url{http://cogcomp.org/page/publication_view/929}.}

\end{abstract}

\section{Introduction}

A multilingual knowledge base (KB) such as DBpedia~\cite{lehmann2015dbpedia}, ConceptNet~\cite{speer2017conceptnet} and Yago~\cite{mahdisoltani2014yago3} stores multiple language-specific knowledge graphs (KGs) that express relations of many concepts and real-world entities.
As each KG thereof is either extracted independently from  monolingual corpora~\cite{lehmann2015dbpedia,mahdisoltani2014yago3} or contributed by speakers of the language~\cite{speer2017conceptnet,mitchell2018never},
it is common for different KGs to constitute complementary knowledge~\cite{bleiholder2009data,bryl2014learning}.
Hence, aligning and synchronizing language-specific KGs support AI systems with more comprehensive commonsense reasoning~\cite{lin2019kagnet,li2019teaching,yeo2018machine},
and benefit various knowledge-driven NLP tasks, including machine translation~\cite{moussallem2018machine}, narrative prediction~\cite{chen2019incorporating} and dialogue agents~\cite{sun2019dream}.

\begin{figure}[!t]
	\centering
	\includegraphics[width=0.87\linewidth]{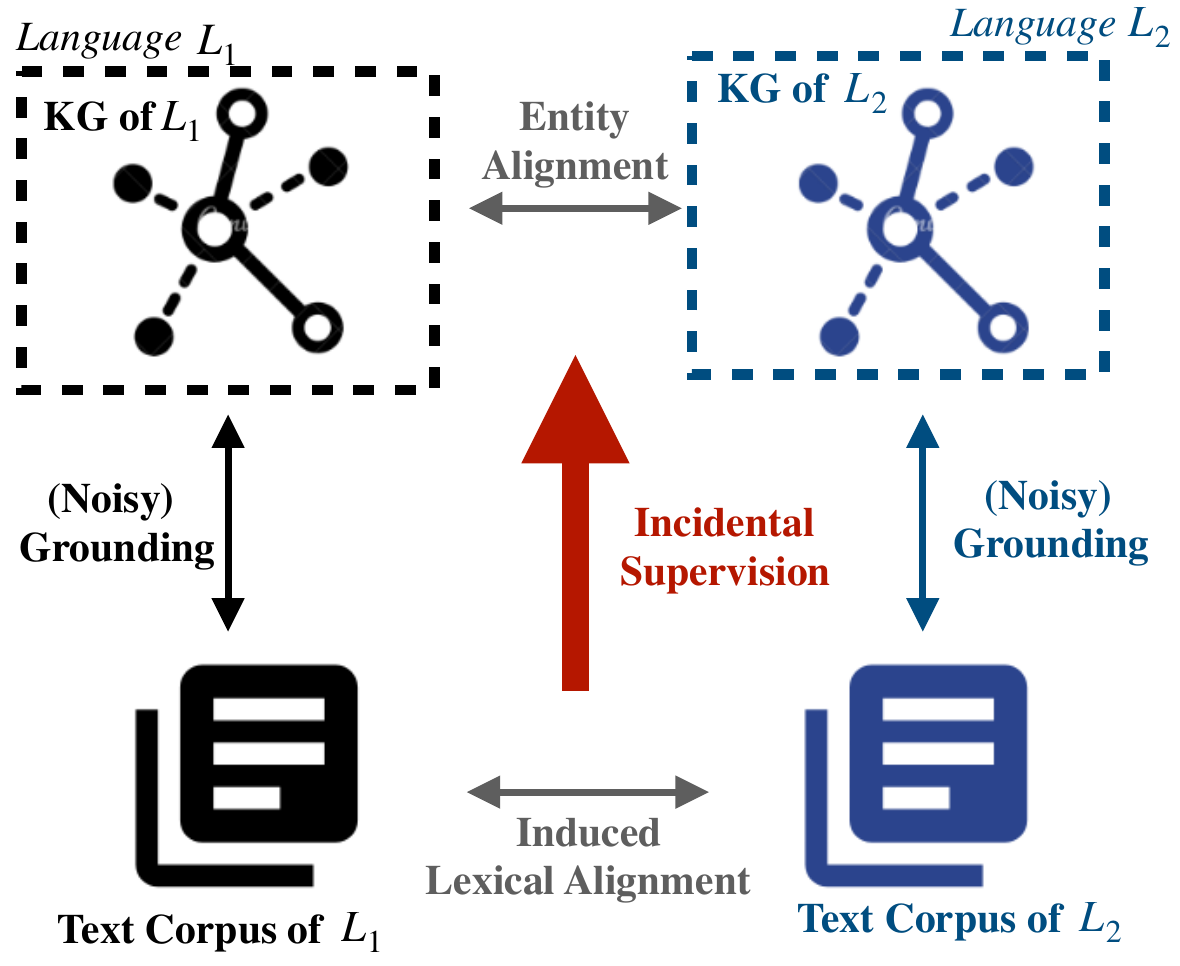}
	\caption{The learning framework of \modelname.}
	\label{fig:arch}
\end{figure}

Learning to align multilingual KGs is a non-trivial task,
as KGs with distinct surface forms, heterogeneous schemata and inconsistent structures easily cause traditional symbolic methods to fall short~\cite{suchanek2011paris,wijaya2013pidgin,jimenez2012large}.
Recently, much attention has been paid to methods based on multilingual KG embeddings~\cite{chen2017multigraph,chen2017akbc,chen2018co,sun2017cross,sun2018bootstrapping,sun2019transedge,zhang2019multi}.
Those methods seek to separately encode the structure of each language-specific KG in an embedding space.
Then, based on some seed entity alignment, the entity counterparts in different KGs can be easily matched via distances or transformations of embedding vectors.
The principle is that entities with relevant neighborhood information can be characterized with similar embedding representations. Such representations particularly are tolerant to the aforementioned heterogeneity of surface forms and schemata in language-specific KGs~\cite{chen2017multigraph,sun2018bootstrapping,sun2020alinet}.

While multilingual KG embeddings provide a general and tractable way to align KGs,
it still remains challenging for related methods to precisely infer the cross-lingual correspondence of entities.
The challenge is that the seed entity alignment, which serves as the essential training data to learn the connection between language-specific KG embeddings, is often limitedly provided in KBs~\cite{chen2018co,sun2018bootstrapping}.
Hence, the lack of supervision often hinders the precision of inferred entity counterparts, and affects even more significantly when KGs scale up and become inconsistent in contents and density~\cite{pujara2017sparsity}.
Several methods also gain auxiliary supervision from profile information of entities, including descriptions~\cite{chen2018co,yang2019aligning,zhang2019multi} and numerical attributes~\cite{sun2017cross,distiawanTrsedya2019}.
However, such profile information is not available in many KGs~\cite{speer2017conceptnet,mitchell2018never,bond2013linking}, therefore causing these methods to be not generally applicable.

Unlike existing models that rely on internal information of KGs, we seek to create embeddings that incorporate both KGs and freely available text corpora, and exploit incidental supervision signals~\cite{roth2017incidental} from text corpora to enhance the alignment learning on KGs (\Cref{fig:arch}).
In this paper, we propose a novel embedding model \modelname (\textbf{J}oint \textbf{E}mbedding Based Entity \textbf{A}lignment with I\textbf{N}cidental \textbf{S}upervision; \includegraphics[height=1em]{fig/1F456.pdf}).
Particularly, \modelname first performs a \emph{grounding process}~\cite{gupta2017entity,upadhyay2018joint} to link entity mentions in each monolingual text corpus to the KG of the same language.
Based on the KGs and grounded text in a pair of languages, \modelname conducts two learning processes, i.e. \emph{embedding learning} and \emph{alignment learning}.
The embedding learning process distributes entities, relations and lexemes of each language in its embedding space,
in which a KG embedding model and a language model for that language are jointly trained.
This process seeks to leverage text contexts to help capture the proximity of entities.
On top of that, alignment learning captures cross-lingual correspondence for entities and lexemes in a self-learning manner~\cite{artetxe2018robust}.
Starting from a small amount of seed entity alignment, this process iteratively induces a transformation between language-specific embedding spaces, and infers more alignment of entity and lexemes at each iteration to improve the learning at the next one.
Moreover, we also employ the closed-form Procrustes solution~\cite{conneau2018word} to strengthen alignment induction within each iteration.
Experimental results on two benchmarks confirm the effectiveness of \modelname in leveraging incidental supervision, leading to significant improvement to entity alignment and drastically outperforming existing methods.

\section{Related Work}\label{sec:related}

We discuss relevant works in four topics.
Each of them has a large body of work which we can only provide as a highly selected summary.

\stitle{Entity alignment.}
Entity alignment in KBs has been a long-standing problem~\cite{shvaiko2011ontology}.
Aside from earlier approaches based on symbolic or schematic similarity of entities~\cite{suchanek2011paris,wijaya2013pidgin,jimenez2012large}, more recent research addresses this task with multilingual KG embeddings.
A representative method of such is MTransE~\cite{chen2017multigraph}.
MTransE jointly learns two model components.
There are a translational embedding model~\cite{bordes2013translating} that distributes the facts in language-specific KGs into separate embeddings, and a transformation-based alignment model that 
maps between entity counterparts across embedding spaces.

Following the general principle of MTransE, later approaches are developed through the following three lines.
One is to incorporate various \emph{embedding learning techniques} for KGs.
Besides translational techniques, some models employ alternative \emph{relation modeling} techniques to encode relation facts, such as circular correlation~\cite{nickel2016holographic}, Hadamard product~\cite{hao2019joie} and recurrent skipping networks~\cite{guo2019learning}.
Others encode entities with \emph{neighborhood aggregation} techniques, including GCN~\cite{wang2018cross,yang2019aligning,cao2019multi,xu2019cross,wu2019jointly}, RGCN~\cite{wu2019relation} and GAT~\cite{zhu2019neighborhood}.
Their benefits are mainly to produce entity representations capturing high-order proximity, so as to better suit the alignment task.
A few works follow the second line to enhance the \emph{alignment learning} with semi-supervised learning techniques.
Representative ones include co-training~\cite{chen2018co}, optimal transport~\cite{pei2019transport} and bootstrapping~\cite{sun2018bootstrapping},
which improve the preciseness of alignment captured with limited supervision.
The third line of research seeks to obtain additional supervision from entity profiles, including descriptions~\cite{chen2018co,yang2019aligning}, attributes~\cite{sun2017cross,distiawanTrsedya2019,pei2019deg,yangcotsae} and KG schemata~\cite{zhang2019multi}.
While those alternative views of entities can effectively bridge the embeddings, the limitation of such methods lies in the unavailability of those views in many KGs~\cite{speer2017conceptnet,mitchell2018never,bond2013linking}.
A survey on the entity alignment problem by~\citet{sun2020benchmark} has provided a more comprehensive summarization of recent advances in these lines.

Our method is mainly related to the third line of research. While instead of leveraging specific intra-KB information, our method introduces supervision signals from text contexts that are freely accessible to almost any KBs with the aid of grounding techniques.
Meanwhile, our paper also follows the second line to improve alignment learning techniques, and couples two mainstream techniques for embedding learning.


\stitle{Joint embeddings of entities and text.}
Fewer efforts have been put to jointly characterize entities and text as embeddings.
\citeauthor{wang2014joint}~\shortcite{wang2014joint} propose to connect a translational embedding of Freebase~\cite{bollacker2008freebase} to a English word embedding based on Wikipedia anchors, therefore providing a joint embedding to enhance link prediction in the KG.
\citeauthor{zhong2015aligning}~\shortcite{zhong2015aligning}  generalize the approach by~\citet{wang2014joint} with distant supervision based on entity descriptions and text corpora.
\citeauthor{toutanova2015representing}~\shortcite{toutanova2015representing} extract dependency paths from sentences and jointly embed them with a KG using DistMult~\cite{yang2015embedding} to support the relation extraction task.
Several other approaches focus on jointly embedding words, entities~\cite{yamada2017learning,newman2018jointly,cao2017bridge,almasian2019word} and entity types~\cite{gupta2017entity} appearing in the same textual contexts without considering relational structure of a KG.
These approaches are employed in monolingual NLP tasks including entity linking~\cite{gupta2017entity,cao2017bridge}, entity abstraction~\cite{newman2018jointly} and factoid QA~\cite{yamada2017learning}.
As they focus on a monolingual and supervised scenario,
they are essentially different from our goal to help cross-lingual KG alignment with incidental supervision from unparalleled corpora.

\stitle{Multilingual word embeddings.}
Our model component of alignment induction from text is closely connected to multilingual word embeddings.
Earlier approaches in this line, regardless of being supervised or weakly supervised, based on seed lexicon~\cite{zou2013bilingual} or parallel corpora~\cite{gouws2015bilbowa},
are systematically summarized in a recent survey \shortcite{ruder2017survey}.
While a number of methods in this line can be employed in our model to gain addition supervision for entity alignment,
we choose to use a combination of Procrustes solution~\cite{conneau2018word} with self-learning to offer precise inference of cross-lingual alignment based on limited seed alignment.
Note that recent contextualized embeddings such M-BERT~\cite{pires2019multilingual} and XLM-R~\cite{conneau2020unsupervised} do not directly suit our problem setting, since contextualization could cause ambiguity to entity representations, therefore impairing the alignment of entities across embedding spaces.

\stitle{Incidental supervision.}
Incidental supervision is a recently introduced learning strategy~\cite{roth2017incidental},
which seeks to retrieve supervision signals from data that are not 
labeled for 
the target task.
This strategy has been applied to tasks including SRL~\cite{he2019incidental}, controversy prediction~\cite{rethmeier2018learning} and dataless classification~\cite{song2015unsupervised}.
To the best of our knowledge, the proposed method here is the first of its kind that incorporates incidental supervision in embedding learning or alignment.

\section{Method}

We hereby begin introducing our method with the formalization of learning resources.

In a \mbox{KB}, $\lang$ denotes the set of languages, and $\lang^2$ 
unordered language pairs.
$G_{L}$ is the language-specific KG of language $L \in \lang$.
$E_{L}$ and $R_{L}$ respectively denote the corresponding vocabularies of entities and relations.
$T = (h,r,t)$ denotes a triple in $G_{L}$ such that $h, t \in E_{L}$ and $r \in R_{L}$.
Boldfaced $\mathbf{h}$, $\mathbf{r}$, $\mathbf{t}$
represent the embedding vectors of head $h$, relation $r$, and tail $t$ respectively. For a language pair
$(L_1, L_2) \in \lang^2$,
$I_E(L_1, L_2)$ denotes a set of entity alignments 
between $L_1$ and $L_2$, such that $e_1\in E_{L_1}$ and $e_2\in E_{L_2}$ for each entity pair $(e_1, e_2)\in I_E(L_1, L_2)$.
Following the convention of previous work~\cite{chen2018co,sun2018bootstrapping,yang2019aligning},
we assume the entity pairs to have a 1-to-1 mapping and it is specified in $I_E(L_1, L_2)$.
This assumption is congruent to the design of mainstream KBs~\cite{lehmann2015dbpedia,mahdisoltani2014yago3} where disambiguation of entities is granted.
Besides the definition of multilingual KGs, we use $D_L$ to denote the text corpus of language $L$.
$D_L$ is a set of documents $\{d_L\}$, where each document $d_L=[w_1, w_2, ..., w_l]$ is a sequence of tokens from the monolingual lexicon $W_L$.
Each token $w_i$ thereof is originally a lexeme, but may also be an entity surface form after the ground process, and we also use boldfaced $\mathbf{w}_i$ to denote its vector.
$I_W(L_1,L_2)$ denotes the seed lexicon between $(L_1,L_2)$, such that $w_1\in W_{L_1}$ and $w_2\in W_{L_2}$ for each lexeme pair $(w_1, w_2)\in I_W(L_1, L_2)$.
Note that $I_W$ 
only include the alignment between lexemes, and may optionally serve as external supervision data.
To be consistent with previous problem settings of entity alignment~\cite{chen2017multigraph,sun2018bootstrapping,yang2019aligning}, $I_W$ is not necessarily provided to training, but is defined to be compatible with the scenarios where it is available.

\modelname addresses entity alignment in three consecutive processes.
(i) A \emph{grounding process} first link entities of each KG to possible mentions of them in the corresponding monolingual corpus,
therefore connecting entities and text tokens of the same language into a shared vocabulary.
(ii) An \emph{embedding learning} process characterizes the KG and text of each language into a separate embedding space.
In this process, we couple both the translational technique~\cite{bordes2013translating,chen2017multigraph,chen2018co} and the neighborhood aggregation technique~\cite{wang2018cross,yang2019aligning}, which are two representative techniques to characterize a KG.
Simultaneously, the monolingual text tokens are encoded with a skip-gram language model~\cite{mikolov2013efficient}.
(iii) On top of the embeddings, starting from a small amount of seed entity alignment and \emph{optional} seed lexicon, the \emph{alignment learning} process iteratively infers more alignment both on KGs and text using self-learning and Procrustes solution~\cite{schonemann1966procrustes}.
The processes of \modelname's learning is consistent to \Cref{fig:arch}.
The rest of this section introduces the technical details of each process.

\subsection{(Noisy) Entity Grounding}

The goal of the grounding process is to combine vocabularies of the KG and the text corpus in each language.
This serves as the premise for the embedding learning process to produce a shared representation scheme for entities, relations and lexemes,
therefore allowing the alignment learning process to leverage supervision signals for both entities and lexemes.
It is noteworthy that, the purpose of entity grounding here is to combine the two data modalities.
Hence, we only expect this process to discover enough entity contexts and offer a higher coverage on entity vocabularies,
while being tolerant to possible noise in entity recognition and linking.
Particularly, we consider two grounding techniques, one using a pre-trained entity discovery and linking (EDL) model, the other based on simple surface form matching (SFM).

\stitle{Pre-trained EDL model.} One technique is to use off-the-shelf EDL models~\cite{khashabi2018cogcompnlp,manning2014stanford}.
A typical model of such sequentially handles the steps of NER to detect entity mentions, and link each mention to candidate entities from the KG based on symbolic and contextual similarity.
Many EDL models are easily trainable on large text corpora with anchors, and offer promising performance of grounding and disambiguation on multiple languages~\cite{sil2018multi}.
In this paper, we do not go into details to the design of EDL models. 
Interested readers are referred to the aforementioned literature.

\stitle{Surface form matching.} Suppose a pre-trained EDL model is not available, then a simpler way of combining data is to match KG surface forms with text.
This can be efficiently done by building a Completion Trie~\cite{hsu2013space} for multi-token surface forms, and conducting a longest prefix matching~\cite{dharmapurikar2006longest} between surface forms and sub-sequences of text tokens.
While this simple technique does not necessarily disambiguate entity mentions, experiments find it sufficient to combine the two modalities and allow supervision signals from induced lexical alignment to propagate to entities.

Once the entity vocabulary $E_L$ and the lexicon $W_L$ of a language are combined, we assume that entity mentions in $D_L$ are properly tokenized as grounded surface forms in $E_L \cap W_L$.
Specifically, we now use $x$ to denote a token in the grounded corpus $D_L$ that can either be an entity $e$ or a lexeme $w$.
Given the combined learning resources for each language, we next describe the processes of embedding learning and alignment learning.

\subsection{Embedding Learning}

The embedding learning process is responsible for capturing the combined KG and text corpus of each language in a shared embedding space $\mathbb{R}^{k}$.
In this process, \modelname jointly learns two model components to respectively encode units of the KG and the text,
among which the overlaps $E_L \cap W_L$ use shared representations.
We hereby describe these two model components in detail.

\subsubsection{KG Embedding}

As discussed in \Cref{sec:related},
previous approaches respectively leverage two forms of embedding learning techniques: (i) \emph{relation modeling}~\cite{chen2017multigraph,sun2018bootstrapping} 
such as vector translations,
circular correlation and Hadamard product seeks to capture relations as an arithmetic operation in the vector space;
(ii) \emph{neighborhood aggregation}~\cite{wang2018cross,yang2019aligning,cao2019multi} employs graph neural networks (GNN) to encode neighborhood contexts for better seizing the proximity of entities.

The KG embedding model proposed in this work couples both forms of techniques. This aims at seizing both relations and entity proximity, two factors that are both beneficial to produce transferable entity embeddings.
To achieve this goal, the encoder first stacks $n$ layers of GCN~\cite{kipf2017gcn} on the KG.
Formally, the $l$-th layer representation $\mathbf{E}^{(l)}$ is computed as

\begin{equation*}
    \mathbf{E}^{(l)}=\phi\left ( D^{-\frac{1}{2}}\tilde{A}D^{-\frac{1}{2}} \mathbf{E}^{(l-1)} \mathbf{M}^{(l-1)} \right ),
\end{equation*}

\noindent
where $D$ is the diagonal degree matrix $D$ of the KG, $\tilde{A}=A+I$ is the sum of the adjacency matrix $A$ and an identity $I$,
and $\mathbf{M}^{(l-1)}$ is a trainable weight matrix.
The raw features of entities $\mathbf{E}^{(0)}$ can be either entity attributes or randomly initialized.
The last layer outputs are regarded as entity embedding representations, i.e. $\mathbf{E}=\mathbf{E}^{(n)}$.

We use $\mathbf{E}_L$ to denote the entity representations of language $L$,
then the following log-softmax loss is optimized to perform relational modeling with translation vectors in the embedding space of $L$:

\begin{equation*}
    S^K_L=-\sum_{T\in G_L}\log\frac{\exp\left (b -f_r(h,t) \right )}{\sum_{\hat{T}\notin G_L}\exp\left (b -f_r(\hat{h},\hat{t}) \right )},
\end{equation*}

\noindent
where $f_r(h,t)=\left \| \mathbf{h}+\mathbf{r}-\mathbf{t}\right \|$ is the plausibility measure of a triple~\cite{bordes2013translating},
$\hat{T}=(\hat{h},r,\hat{t})$ is a Bernoulli negative-sampled triple~\cite{wang2014knowledge} created by substituting either head or tail entities $h$ or $t$ in $T=(h,r,t)$.
$b$ is a positive bias to adjust the scale of the plausibility measure. 
All the entity representations optimized in $S^K_L$ are from $\mathbf{E}_L$.
Note that the reason for us to choose the translational technique over other relation modeling techniques is due to this technique being more robust in cases where KG structures are sparser~\cite{pujara2017sparsity}.

\subsubsection{Text Embedding}

In addition to the KG embedding,
the text embedding seeks to leverage the contextual information of free text to help the embedding better capture the proximity of entities
This model employs the continuous skip-gram language model, 
which is inline with a number of word embedding methods~\cite{mikolov2013efficient,bojanowski2017enriching,conneau2018word}, and is realized by optimizing the following log-softmax loss:

\begin{equation*}
    S^T_L=-\sum_{x\in E_L\cup W_L} \sum_{x_c\in C_{x,D_L}}\log\frac{\exp \left ( d(x, {x}_c) \right )}{\sum_{x_n}\exp \left ( d(x, {x}_n) \right )}.
\end{equation*}

\noindent
The text context $C_{x,D_L}$ thereof is the set of tokens that surround a token $x$ in the entity-grounded corpus $D_L$, $d$ denotes the $l_2$ distance, and $x_n$ denotes a randomly sampled token in $E_L\cup W_L$.

\subsubsection{Embedding Learning Objective}

For each language $L\in \lang$, the goal of embedding learning is to optimize the joint loss 
$$S^E_L=S^K_L+S^T_L.$$
As mentioned, the grounded entity surface forms in $E_L\cap W_L$ use shared representations in both model components, hence are optimized with both $S^K_L$ and $S^T_L$.
The rest lexeme, relation and entity representations are optimized alternately by either component.
In both model components, the number of negative samples of triples and tokens are both adjustable hyperparameters.

It is noteworthy that, both model components may choose alternative techniques, including other KG encoders such as GAT~\cite{velivckovic2018gat}, multi-channel GCN~\cite{cao2019multi} and gated GNN~\cite{sun2020alinet}, and text embeddings such as GloVe~\cite{pennington2014glove}.
As experimenting with different embedding techniques is not a main contribution of this work, we leave them as future work.
Specifically, contextualized text representations~\cite{peters2018deep,devlin2019bert} cannot directly apply, as contextualization will cause ambiguity to token representations that hinder the match of embeddings.

\subsection{Alignment Learning}

Once the KG and text units of each language are captured in a shared embedding,
the alignment learning process therefore bridges the alignment between each pair of embeddings.
This process seeks to exploit additional alignment labels from text embeddings,
and use those to help the alignment of entities.
Different from the majority of methods in \Cref{sec:related} that jointly learn embeddings and alignment, the alignment learning process in \modelname is a retrofitting process~\cite{shi2019retrofit,faruqui2015retrofitting}.
Hence, the embedding of each language is fixed and does not require duplicate training for different language pairs~\cite{chen2017multigraph,sun2017cross}.

Given a pair of languages $(L_i, L_j)\in \lang^2$, the objective of alignment learning is to induce a transformation $\mathbf{M}_{ij}\in \mathbb{R}^{k\times k}$ between the two embedding spaces. The following loss is minimized

\begin{equation*}
    S^A_{L_i,L_j}=\sum_{(x_i,x_j)\in I(L_i,L_j)}\left \| \mathbf{M}_{ij}\mathbf{x}_i - \mathbf{x}_j \right \|_2,
\end{equation*}

\noindent
in which $I(L_i,L_j) = I_E(L_i,L_j) \cup I_W(L_i,L_j)$, and the word seed lexicon $I_W$ is considered additional supervision data that are optionally provided.
Each $\mathbf{x}_i$ ($\mathbf{x}_j$) denotes a fixed representation of either an entity or a lexeme of $L_i$ ($L_j$).

Starting from a small amount of seed alignment in $I(L_i,L_j)$, \modelname conducts an iterative self-learning process to exploit more alignment labels for both entities and lexemes to improve the learning of $\mathbf{M}_{ij}$.
In each iteration, we follow~\citeauthor{conneau2018word}~\shortcite{conneau2018word} to induce a Procrustes solution for $\mathbf{M}_{ij}$.
To propose new alignment labels, the self-learning technique in \modelname deploys a mutual nearest neighbor (NN) constraint, which requires a suggested pair of matched items to appear in the NN of each other.
More specifically, define $\mathcal{N}^K_{L_i}(\mathbf{x})$ as the $K$-NN of vector $\mathbf{x}$ in the embedding space of $L_i$, this constraint requires a proposed match $(x_i,x_j)$ to be inserted into $I$ only if $\mathbf{M}_{ij}\mathbf{x}_i$ is in $\mathcal{N}^1_{L_j}(\mathbf{x}_j)$, and $\mathbf{x}_j$ mutually appears in $\mathcal{N}^1_{L_j}(\mathbf{M}_{ij}\mathbf{x}_i)$.
Besides, we also require $(x_i,x_j)$ to be of the same type, i.e. both being entities or being lexemes.
Particularly, we only select entities that have not been aligned in $I$ to form the newly-proposed $(x_i,x_j)$.
This respects the 1-to-1 matching constraint of entities being defined at the beginning of this section, and effectively reduces the candidate space after each iteration of self-learning.
Meanwhile, 1-to-1 matching is not required for lexemes.
To mitigate hubness, 
we also follow~\citeauthor{conneau2018word}~\shortcite{conneau2018word} to employ the Cross-domain Similarity Local Scaling (CSLS) measure.

After the iteration, the newly proposed alignment labels are inserted to $I$ to enhance the learning at the next iteration.
The iterative self-learning is stopped once the number of proposed entity alignment in an iteration is below certain quantity (e.g. $1\%$ of $|E_{L_i}|$).
With more and more matched entities and lexemes being exploited within each iteration,
a better $\mathbf{M}_{ij}$ is induced, whereas the lexical alignment naturally serve as incidental supervision signals for entity alignment.

{
\begin{table*}[t]
\setlength\tabcolsep{2pt}
\centering
\scriptsize
\begin{tabular}{c|ccc|ccc|ccc|ccc|ccc}
\bhline
Settings&\multicolumn{3}{c|}{DBP15k$_{En-Fr}$}&\multicolumn{3}{c|}{DBP15k$_{En-Zh}$}&\multicolumn{3}{c|}{DBP15k$_{En-Ja}$}&\multicolumn{3}{c|}{WK3l60k$_{En-Fr}$}&\multicolumn{3}{c}{WK3l60k$_{En-De}$}\\
\hline
Metrics&$\hitsone$&$\hitsten$&$\mrr$&$\hitsone$&$\hitsten$&$\mrr$&$\hitsone$&$\hitsten$&$\mrr$&$\hitsone$&$\hitsfive$&$\mrr$&$\hitsone$&$\hitsfive$&$\mrr$\\
\bhline
MTransE \cite{chen2017multigraph}$\dagger\ddagger$&0.224&0.556&0.335&0.308&0.614&0.364&0.279&0.575&0.349&0.140&0.203&0.177&0.034&0.101&0.072\\
GCN-Align \cite{wang2018cross}$\ddagger$&0.373&0.745&0.532&0.413&0.744&0.549&0.399&0.745&0.546&0.215&0.378&0.293&0.138&0.246&0.190\\
AlignE \cite{sun2018bootstrapping}$\dagger$&0.481&0.824&0.599&0.472&0.792&0.581&0.448&0.789&0.563&$--$&$--$&$--$&$--$&$--$&$--$\\
GCN-JE \cite{wu2019jointly}&0.483&0.778&$--$&0.459&0.729&$--$&0.466&0.746&$--$&$--$&$--$&$--$&$--$&$--$&$--$\\
RotatE \cite{sun2019rotate}$\dagger$&0.345&0.738&0.476&0.485&0.788&0.589&0.442&0.761&0.550&$--$&$--$&$--$&$--$&$--$&$--$\\
KECG \cite{li2019semi}&0.486&0.851&0.610&0.478&0.835&0.598&0.490&0.844&0.610&$--$&$--$&$--$&$--$&$--$&$--$\\
MuGCN \cite{cao2019multi}$\dagger$&0.495&0.870&0.621&0.494&0.844&0.611&0.501&0.857&0.621&$--$&$--$&$--$&$--$&$--$&$--$\\
RSN \cite{guo2019learning}$\dagger$&0.516&0.768&0.605&0.508&0.745&0.591&0.507&0.737&0.590&$--$&$--$&$--$&$--$&$--$&$--$\\
GMN \cite{xu2019cross}&0.596&0.876&0.679&0.433&0.681&0.479&0.465&0.728&0.580&$--$&$--$&$--$&$--$&$--$&$--$\\
AliNet \cite{sun2020alinet}$\dagger$&0.552&0.852&0.657&0.539&0.826&0.628&0.549&0.831&0.645&$--$&$--$&$--$&$--$&$--$&$--$\\
\hline
JAPE \cite{sun2017cross}$\dagger\ddagger$&0.324&0.667&0.430&0.412&0.745&0.490&0.363&0.685&0.476&0.169&0.354&0.271&0.147&0.239&0.192\\
SEA \cite{pei2019deg}$\dagger$&0.400&0.797&0.533&0.424&0.796&0.548&0.385&0.783&0.518&$--$&$--$&$--$&$--$&$--$&$--$\\
HMAN \cite{yang2019aligning}&0.543&0.867&$--$&0.537&0.834&$--$&0.565&0.866&$--$&$--$&$--$&$--$&$--$&$--$&$--$\\
\hline
BootEA \cite{sun2018bootstrapping}$\dagger\ddagger$&0.653&0.874&0.731&0.629&0.847&0.703&0.622&0.854&0.701&0.333&0.511&0.425&0.233&0.393&0.316\\
KDCoE \cite{chen2018co}&$--$&$--$&$--$&$--$&$--$&$--$&$--$&$--$&$--$&\textbf{0.483}&\textbf{0.569}&0.496&0.335&0.380&0.339\\
MMR \cite{shi2019modeling}&0.635&0.878&$--$&0.647&0.858&$--$&0.623&0.847&$--$&$--$&$--$&$--$&$--$&$--$&$--$\\
NAEA \cite{zhu2019neighborhood}&0.673&0.894&0.752&0.650&0.867&0.720&0.641&0.873&0.718&$--$&$--$&$--$&$--$&$--$&$--$\\
OTEA \cite{pei2019transport}$\ddagger$&$--$&$--$&$--$&$--$&$--$&$--$&$--$&$--$&$--$&0.361&0.541&0.447&0.270&0.440&0.352\\
\hline
\modelname-SFM&0.766&0.939&0.814&0.713&0.885&0.773&0.723&0.913&0.793&0.463&0.558&\textbf{0.538}&\textbf{0.337}&\textbf{0.450}&\textbf{0.412}\\
\modelname-EDL&\textbf{0.769}&\textbf{0.940}&\textbf{0.827}&\textbf{0.719}&\textbf{0.895}&\textbf{0.791}&\textbf{0.737}&\textbf{0.914}&\textbf{0.798}&0.451&0.544&0.529&0.312&0.431&0.390\\
\bhline
\end{tabular}
\caption{Entity alignment results. Baselines are separated in accord with the three groups described in Section~\ref{sec:exp_set}. $\dagger$ indicates results obtained from \cite{sun2020alinet}, and $\ddagger$ indicates those from \cite{pei2019transport}. Results of KECG, GCN-JE, MMR, HMAN, KDCoE and NAEA are from original papers. Hyphens denote not available.
MRR were not reported by GCN-JE, MMR and HMAN. Top results (incl. w/ and w/o seed lexicon) are boldfaced.
Note that results by GCN-JE, GMN and HMAN are reported only for the versions where the extra cross-lingual alignment information (such as machine translation) is removed, so as to conduct fair comparison with all the rest models that are trained using only the alignment labels in the benchmark training sets.
}\label{tbl:entity}
\end{table*}
}


After the alignment learning process, given a query $(e_i,?e_j)$ to find the counterpart entity of $e_i\in E_{L_i}$ from $E_{L_j}$,
the answer $e_j$ is predicted as the $1$-NN entity after applying $\mathbf{M}_{ij}$ to transform $\mathbf{e}_i$, denoted $\{e_j\}=\mathcal{N}^1_{E_{L_i}}(\mathbf{M}_{ij}\mathbf{e}_i)$.
The inference phase by default also adopts CSLS as the distance measure, which is consistent with the default setting of recent works~\cite{sun2019transedge,sun2020alinet}.
\section{Experiment}\label{sec:exp}

In this section, we evaluate \modelname on two benchmark datasets for cross-lingual entity alignment, and compare against a wide selection of recent baseline methods.
We also provide detailed ablation study on model components of \modelname.

\subsection{Experimental Settings}\label{sec:exp_set}

\stitle{Datasets.} Experiments are conducted on DBP15k~\cite{sun2017cross} and WK3l60k~\cite{chen2018co} that are widely used benchmarks on the studied task.
DBP15k contains four language-specific KGs that are respectively extracted from English (En), Chinese (Zh), French (Fr) and Japanese (Ja) DBpedia~\cite{lehmann2015dbpedia}, each of which contains around 65k-106k entities. 
Three sets of 15k alignment labels are constructed to align entities between each of the other three languages and English.
WK3l60k contains larger KGs with around 57k to 65k entities in En, Fr and German (De) KGs, and around 55k reference entity alignment for En-Fr and En-De settings.
Dataset statistics are given in Appendix~\Cref{sup:stats}~\cite{chen2021cross}.

We also use the text of Wikipedia dumps (dated Jan 1, 2019) in the five participating languages in training.
For Chinese and Japanese corpora thereof, we obtain the segmented versions respectively from PKUSEG~\cite{pkuseg} and MeCab~\cite{kudo2006mecab}.


\stitle{Baseline methods.}
We compare with a wide selection of recent approaches for entity alignment on multilingual KGs.
The baseline methods include (i) those employing different structure embedding techniques, namely 
MTransE~\cite{chen2017multigraph},  GCN-Align~\cite{wang2018cross}, AlignE~\cite{sun2018bootstrapping}, 
GCN-JE~\cite{wu2019jointly},
KECG~\cite{li2019semi},
MuGCN~\cite{cao2019multi}, RotatE~\cite{sun2019rotate}, 
RSN~\cite{guo2019learning} and AliNet~\cite{sun2020alinet}; 
(ii) methods that incorporate auxiliary information of entities, namely JAPE~\cite{sun2017cross}, SEA~\cite{pei2019deg}, GMN~\cite{xu2019cross} and HMAN~\cite{yang2019aligning};
(iii) semi-supervised alignment learning methods, including BootEA~\cite{sun2018bootstrapping}, KDCoE~\cite{chen2018co}, MMR~\cite{shi2019modeling}, NAEA~\cite{zhu2019neighborhood} and OTEA~\cite{pei2019transport}.
Descriptions of these methods are given in Appendix \Cref{sup:base}~\cite{chen2021cross}.

Note that some works have allowed to incorporate extra cross-lingual signals such as machine translation in training, or using pre-aligned word embeddings to delimit candidate spaces~\cite{wu2019relation,wu2019jointly,xu2019cross}. For example,~\citet{wu2019relation,wu2019jointly} used Google Translate to translate surface forms of entities in all other languages to English, and initialize the entity embeddings in their model with pre-trained word embedding of translated entity names. 
Results for these models are reported for the versions where the extra cross-lingual alignment information
is removed so as to conduct fair comparison with all the rest models that are trained from scratch and using the same alignment labels in the benchmark datasets.
This also necessarily prevents potential leakage of testing data to training~\cite{liu2020exploring}, considering that  training a comprehensive NMT system may have subsumed many of the testing data in the entity alignment benchmarks.


\stitle{Evaluation protocols.} 
The use of the datasets are consistent with previous studies of the baseline methods.
On each language pair in DBP15k, around 30\% of seed alignment is used for training, the rest for testing.
On WK3l60k, 20\% of seed alignment on En-Fr and En-De settings is respectively used for training.
Following the convention, we calculate several ranking metrics on test cases, including the accuracy $\hitsone$, the proportion of cases that are ranked no larger than $p$ $\hitsk$, and mean reciprocal rank $\mrr$.
Note that to align with the results in previous studies~\cite{sun2020alinet,pei2019transport}, $p$ is set to 10 on DBP15k and 5 on WK3l60k.
All metrics are preferred higher to indicate better performance.

\stitle{Model Configurations.}
We use AMSGrad~\citep{reddi2018convergence} to optimize the training losses of the embedding learning process, for which we set the learning rate $\alpha$ to 0.001, the exponential decay rates $\beta_1$ and $\beta_2$ to 0.9 and 0.999, and batch sizes to 512 for both $S^K_L$ and $S^T_L$.
Trainable parameters are initialized using Xavier initialization~\cite{glorot2010understanding}.
The dimension $k$ is set to 300, which is often used for bilingual word embedding models trained on Wikipedia corpora~\cite{conneau2018word,gouws2015bilbowa}, considering that the vocabulary sizes and training data density here are relatively close to those models.
The number of GCN layers is set to 2.
We set negative sample sizes of triples and text contexts to 5, the text context width to be 10 and the bias $b$ in $S^K_L$ to be 2.
More implementation details are in Appendix \Cref{sup:hyper}~\cite{chen2021cross}.
Specifically, we evaluate variants of \modelname by adjusting two technical details.
First, for the grounding process, aside from the simple surface form matching (marked with SFM), we also explore with 
the off-the-shelf Wikification-based EDL model (\citeauthor{upadhyay2018joint},~\citeyear{upadhyay2018joint}, marked with EDL).
A grounding performance estimation is given in \Cref{sup:grounding}.
In addition, we consider both CSLS and $l_2$ in inference.

\subsection{Results}
We report the entity alignment results in \Cref{tbl:entity}.

Considering the baseline results on DBP15k, we can see that the simplest variant of \modelname using SFM-based grounding has consistently outperformed all baselines on three cross-lingual settings.
Particularly, it leads to 17.0-17.4\% of absolute improvement in $\hitsone$ over the best structure-based baseline, 14.0-22.3\% over the best entity profile based one, and 6.30-9.30\% over the best semi-supervised one.
This shows that while \modelname preserves the key merit of a semi-supervised entity alignment method,
and effectively enhances the alignment of KGs by exploiting incidental supervision signals from unaligned text corpora.
Considering different grounding techniques, we observe that SFM variants often perform closely to EDL ones.
This indicates that simple SFM is enough to combine KG and text corpora for \modelname's embedding learning without EDL-related resources.
The results on Wk3l60k generally exhibit similar observations. 
In comparison to KDCoE that leverages strong but expensive supervision data of entity descriptions in co-training, \modelname  offers comparable performance based on very accessible resources.

In general, the experiments here show that \modelname promisingly improves SOTA performance for entity alignment, with only the need for unparalleled free text and no need for additional labels.

{
\begin{table}[t]
\setlength\tabcolsep{1pt}
\centering
\footnotesize
\begin{tabular}{l|ccc|ccc}
\bhline
Setting&\multicolumn{3}{c|}{DBP15k$_{En-Fr}$}&\multicolumn{3}{c}{DBP15k$_{En-Ja}$}\\
\hline
Metrics&$\hitsone$&$\hitsten$&$\mrr$&$\hitsone$&$\hitsten$&$\mrr$\\
\bhline
\modelname-SFM&0.766&0.939&0.814&0.723&0.913&0.793\\
\hline
---w/o Self-learning&0.628&0.845&0.720&0.622&0.835&0.728\\
---w/o GCN&0.742&0.913&0.809&0.709&0.905&0.789\\
---w/o Text&0.725&0.891&0.786&0.681&0.857&0.761\\
---w/o KG&0.699&0.872&0.771&0.635&0.819&0.706\\
---w/o CSLS&0.697&0.905&0.762&0.687&0.893&0.768\\
---w/ seed lexicons&0.788&0.947&0.848&0.738&0.931&0.803\\
\bhline
\end{tabular}
\caption{The ablation study results for components of \modelname based on DBP15k$_{En-Fr}$ and DBP15k$_{En-Ja}$. Note that the additional seed lexicon is not used in the main experiment, and is not obligatory. The last row in this table is only to show the effectiveness of leveraging available supervision data on lexemes.}\label{tbl:ablation}
\end{table}
}

{
\begin{table}[t]
\setlength\tabcolsep{2pt}
\centering
\footnotesize
\begin{tabular}{c|cc|cc}
\bhline
Language&\multicolumn{2}{c|}{SFM}&\multicolumn{2}{c}{EDL}\\
\hline
Estimation&Coverage&Avg match&Coverage&Avg match\\
\hline
En&0.982&1,268&0.933&1,367\\
Fr&0.987&295&0.926&929\\
Zh&0.855&141&0.774&348\\
Ja&0.982&159&0.797&881\\
De&0.981&297&0.951&1,092\\
\bhline
\end{tabular}
\caption{Estimated vocabulary coverage and average match per entity on each of the five language-specific Wikipedia corpora. }\label{tbl:statgrounding}
\end{table}
}

\subsection{Ablation Study}

In \Cref{tbl:ablation} we report an ablation study for \modelname-SFM based on DBP15k, so as to understand the importance of each incorporated technique.

From the results, we observe that self-learning is the most important factor.
The removal of it can lead to a drop of 10.1-13.8\% in $\hitsone$, as well as drastic drop of other metrics.
This also explains why semi-supervised baselines (group 3) typically perform better than others.
However, even with self-learning, the removal of text can lead to $\hitsone$ drop of 2.4\% on En-Fr and 4.2\% on En-Ja.
This shows that context information \modelname retrieves from free text effectively infers the match of entities.
On the other hand, the structure encoding of KGs is more important than textual contexts,
as it causes higher performance drops of 6.7-8.8\% in $\hitsone$ by removing KGs.
Note that the model without KG learns entity embeddings solely based on free text. Its results show that context information from text alone can provide a strong starting point from which incorporating KGs can further enhance its performance.
Employing GCN leads to relatively slight performance gain, as joint learning the relation model and the language model can satisfyingly capture entity proximity.
Changing the distance metric to $l_2$ also leads to 3.6-6.9\% of decrease in $\hitsone$.
This shows CSLS's ability to handle hubness and isolation is also important for similarity inference in the dense embedding space for the metric words and entities.
Hence, this metric is also recommended by recent work~\cite{sun2020alinet,sun2019transedge,zhang2019multi}. 
In addition, if we introduce additional 5k seed lexicon (with only word alignment information, not including any entity alignment) provided by~\citeauthor{conneau2018word}~\shortcite{conneau2018word} for each language pair, it leads to additional improvement of 1.5-2.2\% in $\hitsone$.
This shows that \modelname effectively leverages available supervision data on lexemes to further enhance entity alignment, although it is not obligatory.

\subsection{Grounding Performance Estimation}\label{sup:grounding}

Due to the lack of ground truths on unlabeled text, it is hard to estimate the precision of entity grounding by the two types of (noisy) grounding techniques.
However, as the requirement of the grounding process is to simply connect two data modalities for training the embeddings, 
we may encourage a technique that handles enough entity mentions and offer a higher coverage on entity vocabularies.
Accordingly, the estimations of these two factors for the two techniques are reported in \Cref{tbl:statgrounding}.
As we can observe that, without considering disambiguation, SFM can overall cover higher proportions of the entity vocabularies, while pre-trained EDL generally discovers more entity mentions for each entity.
However, both techniques are sufficient to support the noisy grounding process and combine two data modalities for embedding learning and alignment induction.
\section{Conclusion}

This paper introduces \modelname for entity alignment.
Different from previous methods that leverage only internal information of KGs, \modelname extends the learning on any text corpora that may contain the KG entities.
For each language, a noisy grounding process first connects both data modalities, followed by an embedding learning process coupling GCN with relational modeling, and an self-learning based alignment process.
Without introducing additional labeled data,
\modelname offers significantly improved performance over SOTA models on benchmarks.
Hence, it shows the effectiveness and feasibility of exploiting incidental supervision from free text for entity alignment.

For future work, aside from experimenting with other embedding learning techniques for KGs and text, we plan to extend \modelname to learn associations on KGs with different specificity~\cite{hao2019joie}.
We also seek to extend the representation scheme in hyperbolic spaces~\cite{nickel2017poincare,chen2019edge} along with the incorporation of hyperbolic lexical embedding techniques~\cite{tifrea2018poincare}, aiming at better capturing the associations for hierarchical ontologies.

\section*{Acknowledgement}
We appreciate the anonymous reviewers for their insightful comments and suggestions.
Also, we would like thank Jennifer Sheffield, Yi Zhang and other members of 
the UPenn Cognitive Computation Group for giving suggestions 
that improved 
the manuscript.

This research is supported in part by the Office of the Director of National Intelligence (ODNI), Intelligence Advanced Research Projects Activity (IARPA), via IARPA Contract No. 2019-19051600006 under the BETTER Program, and by Contract FA8750-19-2-1004 with the US Defense Advanced Research Projects Agency (DARPA). The views expressed are those of the authors and do not reflect the official policy or position of the Department of Defense or the U.S. Government.

\bibliography{acl2019}

\begin{thebibliography}{92}
\expandafter\ifx\csname natexlab\endcsname\relax\def\natexlab#1{#1}\fi

\bibitem[{Almasian et~al.(2019)Almasian, Spitz, and Gertz}]{almasian2019word}
Satya Almasian, Andreas Spitz, and Michael Gertz. 2019.
\newblock Word embeddings for entity-annotated texts.
\newblock In \emph{Proceedings of the European Conference on Information
  Retrieval (ECIR)}.

\bibitem[{Artetxe et~al.(2018)Artetxe, Labaka, and Agirre}]{artetxe2018robust}
Mikel Artetxe, Gorka Labaka, and Eneko Agirre. 2018.
\newblock A robust self-learning method for fully unsupervised cross-lingual
  mappings of word embeddings.
\newblock In \emph{Proceedings of the Annual Meeting of Associations for
  Computational Linguistics (ACL)}.

\bibitem[{Bleiholder and Naumann(2009)}]{bleiholder2009data}
Jens Bleiholder and Felix Naumann. 2009.
\newblock Data fusion.
\newblock \emph{ACM Computing Surveys}, 41(1):1--41.

\bibitem[{Bojanowski et~al.(2017)Bojanowski, Grave, Joulin, and
  Mikolov}]{bojanowski2017enriching}
Piotr Bojanowski, Edouard Grave, Armand Joulin, and Tomas Mikolov. 2017.
\newblock Enriching word vectors with subword information.
\newblock \emph{Transactions of the Association for Computational Linguistics
  (TACL)}, 5.

\bibitem[{Bollacker et~al.(2008)Bollacker, Evans, Paritosh, Sturge, and
  Taylor}]{bollacker2008freebase}
Kurt Bollacker, Colin Evans, Praveen Paritosh, Tim Sturge, and Jamie Taylor.
  2008.
\newblock Freebase: a collaboratively created graph database for structuring
  human knowledge.
\newblock In \emph{Proceedings of ACM SIGMOD International Conference on
  Management of Data (SIGMOD)}.

\bibitem[{Bond and Foster(2013)}]{bond2013linking}
Francis Bond and Ryan Foster. 2013.
\newblock Linking and extending an open multilingual {Wordnet}.
\newblock In \emph{Proceedings of the Annual Meeting of Associations for
  Computational Linguistics (ACL)}.

\bibitem[{Bordes et~al.(2013)Bordes, Usunier, Garcia-Duran, Weston, and
  Yakhnenko}]{bordes2013translating}
Antoine Bordes, Nicolas Usunier, Alberto Garcia-Duran, Jason Weston, and Oksana
  Yakhnenko. 2013.
\newblock Translating embeddings for modeling multi-relational data.
\newblock In \emph{Advances in Neural Information Processing Systems (NIPS)}.

\bibitem[{Bryl and Bizer(2014)}]{bryl2014learning}
Volha Bryl and Christian Bizer. 2014.
\newblock Learning conflict resolution strategies for cross-language wikipedia
  data fusion.
\newblock In \emph{Proceedings of the World Wide Web Conference (WWW)}, pages
  1129--1134.

\bibitem[{Cao et~al.(2017)Cao, Huang, Ji, Chen, and Li}]{cao2017bridge}
Yixin Cao, Lifu Huang, Heng Ji, Xu~Chen, and Juanzi Li. 2017.
\newblock Bridge text and knowledge by learning multi-prototype entity mention
  embedding.
\newblock In \emph{Proceedings of the Annual Meeting of Associations for
  Computational Linguistics (ACL)}.

\bibitem[{Cao et~al.(2019)Cao, Liu, Li, Li, and Chua}]{cao2019multi}
Yixin Cao, Zhiyuan Liu, Chengjiang Li, Juanzi Li, and Tat-Seng Chua. 2019.
\newblock Multi-channel graph neural network for entity alignment.
\newblock In \emph{Proceedings of the Annual Meeting of Associations for
  Computational Linguistics (ACL)}, pages 1452--1461.

\bibitem[{Chen et~al.(2019)Chen, Chen, and Yu}]{chen2019incorporating}
Jiaao Chen, Jianshu Chen, and Zhou Yu. 2019.
\newblock Incorporating structured commonsense knowledge in story completion.
\newblock In \emph{Proceedings of AAAI Conference on Artificial Intelligence
  (AAAI)}.

\bibitem[{Chen and Quirk(2019)}]{chen2019edge}
Muhao Chen and Chris Quirk. 2019.
\newblock Embedding edge-attributed relational hierarchies.
\newblock In \emph{Proceedings of the Annual International ACM SIGIR Conference
  on Research and Development in Information Retrieval (SIGIR)}.

\bibitem[{Chen et~al.(2021)Chen, Shi, Zhou, and Roth}]{chen2021cross}
Muhao Chen, Weijia Shi, Ben Zhou, and Dan Roth. 2021.
\newblock Cross-lingual entity alignment with incidental supervision.
\newblock \emph{arXiv preprint arXiv:2005.00171v2 (with Appendices)}.

\bibitem[{Chen et~al.(2018)Chen, Tian, Chang, Skiena, and Zaniolo}]{chen2018co}
Muhao Chen, Yingtao Tian, Kai-Wei Chang, Steven Skiena, and Carlo Zaniolo.
  2018.
\newblock Co-training embeddings of knowledge graphs and entity descriptions
  for cross-lingual entity alignment.
\newblock In \emph{Proceedings of the International Joint Conference on
  Artificial Intelligence (IJCAI)}.

\bibitem[{Chen et~al.(2017{\natexlab{a}})Chen, Tian, Yang, and
  Zaniolo}]{chen2017multigraph}
Muhao Chen, Yingtao Tian, Mohan Yang, and Carlo Zaniolo. 2017{\natexlab{a}}.
\newblock Multilingual knowledge graph embeddings for cross-lingual knowledge
  alignment.
\newblock In \emph{Proceedings of the International Joint Conference on
  Artificial Intelligence (IJCAI)}.

\bibitem[{Chen et~al.(2017{\natexlab{b}})Chen, Zhou et~al.}]{chen2017akbc}
Muhao Chen, Tao Zhou, et~al. 2017{\natexlab{b}}.
\newblock Multi-graph affinity embeddings for multilingual knowledge graphs.
\newblock In \emph{Automated Knowledge Base Construction (AKBC)}.

\bibitem[{Conneau et~al.(2020)Conneau, Khandelwal, Goyal, Chaudhary, Wenzek,
  Guzm{\'a}n, Grave, Ott, Zettlemoyer, and Stoyanov}]{conneau2020unsupervised}
Alexis Conneau, Kartikay Khandelwal, Naman Goyal, Vishrav Chaudhary, Guillaume
  Wenzek, Francisco Guzm{\'a}n, Edouard Grave, Myle Ott, Luke Zettlemoyer, and
  Veselin Stoyanov. 2020.
\newblock Unsupervised cross-lingual representation learning at scale.
\newblock In \emph{Proceedings of the Annual Meeting of Associations for
  Computational Linguistics (ACL)}.

\bibitem[{Conneau et~al.(2018)Conneau, Lample, Ranzato, Denoyer, and
  Jégou}]{conneau2018word}
Alexis Conneau, Guillaume Lample, Marc'Aurelio Ranzato, Ludovic Denoyer, and
  Hervé Jégou. 2018.
\newblock Word translation without parallel data.
\newblock In \emph{International Conference on Learning Representations
  (ICLR)}.

\bibitem[{Devlin et~al.(2019)Devlin, Chang, Lee, and
  Toutanova}]{devlin2019bert}
Jacob Devlin, Ming-Wei Chang, Kenton Lee, and Kristina Toutanova. 2019.
\newblock Bert: Pre-training of deep bidirectional transformers for language
  understanding.
\newblock In \emph{Proceedings of the Conference of the North American Chapter
  of the Association for Computational Linguistics: Human Language Technologies
  (NAACL-HLT)}.

\bibitem[{Dharmapurikar et~al.(2006)Dharmapurikar, Krishnamurthy, and
  Taylor}]{dharmapurikar2006longest}
Sarang Dharmapurikar, Praveen Krishnamurthy, and David~E Taylor. 2006.
\newblock Longest prefix matching using bloom filters.
\newblock \emph{IEEE/ACM Transactions on Networking}, 14(2):397--409.

\bibitem[{Faruqui et~al.(2015)Faruqui, Dodge, Jauhar, Dyer, Hovy, and
  Smith}]{faruqui2015retrofitting}
Manaal Faruqui, Jesse Dodge, Sujay~Kumar Jauhar, Chris Dyer, Eduard Hovy, and
  Noah~A Smith. 2015.
\newblock Retrofitting word vectors to semantic lexicons.
\newblock In \emph{Proceedings of the Conference of the North American Chapter
  of the Association for Computational Linguistics: Human Language Technologies
  (NAACL-HLT)}.

\bibitem[{Glorot and Bengio(2010)}]{glorot2010understanding}
Xavier Glorot and Yoshua Bengio. 2010.
\newblock Understanding the difficulty of training deep feedforward neural
  networks.
\newblock In \emph{Proceedings of the International Conference on Artificial
  Intelligence and Statistics (AISTATS)}.

\bibitem[{Gouws et~al.(2015)Gouws, Bengio et~al.}]{gouws2015bilbowa}
Stephan Gouws, Yoshua Bengio, et~al. 2015.
\newblock Bilbowa: Fast bilingual distributed representations without word
  alignments.
\newblock In \emph{Proceedings of the International Conference on Machine
  Learning (ICML)}.

\bibitem[{Guo et~al.(2019)Guo, Sun, and Hu}]{guo2019learning}
Lingbing Guo, Zequn Sun, and Wei Hu. 2019.
\newblock Learning to exploit long-term relational dependencies in knowledge
  graphs.
\newblock In \emph{Proceedings of the International Conference on Machine
  Learning (ICML)}.

\bibitem[{Gupta et~al.(2017)Gupta, Singh, and Roth}]{gupta2017entity}
Nitish Gupta, Sameer Singh, and Dan Roth. 2017.
\newblock Entity linking via joint encoding of types, descriptions, and
  context.
\newblock In \emph{Proceedings of the Conference on Empirical Methods in
  Natural Language Processing (EMNLP)}.

\bibitem[{Hao et~al.(2019)Hao, Chen, Yu, Sun, and Wang}]{hao2019joie}
Junheng Hao, Muhao Chen, Wenchao Yu, Yizhou Sun, and Wei Wang. 2019.
\newblock Universal representationlearning of knowledge bases by jointly
  embedding instances and ontological concepts.
\newblock In \emph{Proceedings of the ACM SIGKDD International Conference on
  Knowledge Discovery and Data Mining (KDD)}.

\bibitem[{He et~al.(2020)He, Ning, and Roth}]{he2019incidental}
Hangfeng He, Qiang Ning, and Dan Roth. 2020.
\newblock Quase: Question-answer driven sentence encoding.
\newblock In \emph{Proceedings of the Annual Meeting of Associations for
  Computational Linguistics (ACL)}.

\bibitem[{Hsu and Ottaviano(2013)}]{hsu2013space}
Bo-June~Paul Hsu and Giuseppe Ottaviano. 2013.
\newblock Space-efficient data structures for top-k completion.
\newblock In \emph{Proceedings of the World Wide Web Conference (WWW)}.

\bibitem[{Jim{\'e}nez-Ruiz et~al.(2012)Jim{\'e}nez-Ruiz, Grau, Zhou, and
  Horrocks}]{jimenez2012large}
Ernesto Jim{\'e}nez-Ruiz, Bernardo~Cuenca Grau, Yujiao Zhou, and Ian Horrocks.
  2012.
\newblock Large-scale interactive ontology matching: Algorithms and
  implementation.
\newblock In \emph{Proceedings of the European Conference on Artificial
  Intelligence (ECAI)}.

\bibitem[{Khashabi et~al.(2018)Khashabi, Sammons, Zhou, Redman,
  Christodoulopoulos, Srikumar, Rizzolo, Ratinov, Luo, Do
  et~al.}]{khashabi2018cogcompnlp}
Daniel Khashabi, Mark Sammons, Ben Zhou, Tom Redman, Christos
  Christodoulopoulos, Vivek Srikumar, Nickolas Rizzolo, Lev Ratinov, Guanheng
  Luo, Quang Do, et~al. 2018.
\newblock Cogcompnlp: Your swiss army knife for nlp.
\newblock In \emph{Proceedings of the International Conference on Language
  Resources and Evaluation (LREC)}.

\bibitem[{Kipf and Welling(2016)}]{kipf2017gcn}
Thomas~N Kipf and Max Welling. 2016.
\newblock Semi-supervised classification with graph convolutional networks.
\newblock In \emph{International Conference on Learning Representations
  (ICLR)}.

\bibitem[{Kudo(2006)}]{kudo2006mecab}
Taku Kudo. 2006.
\newblock Mecab: Yet another part-of-speech and morphological analyzer.
\newblock \emph{http://mecab.sourceforge.jp}.

\bibitem[{Lehmann et~al.(2015)Lehmann, Isele, Jakob, Jentzsch, Kontokostas,
  Mendes, Hellmann, Morsey, Van~Kleef, Auer et~al.}]{lehmann2015dbpedia}
Jens Lehmann, Robert Isele, Max Jakob, Anja Jentzsch, Dimitris Kontokostas,
  Pablo~N Mendes, Sebastian Hellmann, Mohamed Morsey, Patrick Van~Kleef,
  S{\"o}ren Auer, et~al. 2015.
\newblock Dbpedia--a large-scale, multilingual knowledge base extracted from
  wikipedia.
\newblock \emph{Semantic Web}, 6(2):167--195.

\bibitem[{Li et~al.(2019{\natexlab{a}})Li, Cao, Hou, Shi, Li, and
  Chua}]{li2019semi}
Chengjiang Li, Yixin Cao, Lei Hou, Jiaxin Shi, Juanzi Li, and Tat-Seng Chua.
  2019{\natexlab{a}}.
\newblock Semi-supervised entity alignment via joint knowledge embedding model
  and cross-graph model.
\newblock In \emph{Proceedings of the 2019 Conference on Empirical Methods in
  Natural Language Processing and the 9th International Joint Conference on
  Natural Language Processing (EMNLP-IJCNLP)}.

\bibitem[{Li et~al.(2019{\natexlab{b}})Li, Chen, and Yu}]{li2019teaching}
Shiyang Li, Jianshu Chen, and Dian Yu. 2019{\natexlab{b}}.
\newblock Teaching pretrained models with commonsense reasoning: A preliminary
  kb-based approach.
\newblock In \emph{Advances in Neural Information Processing Systems
  (NeurIPS)}.

\bibitem[{Lin et~al.(2019)Lin, Chen, Chen, and Ren}]{lin2019kagnet}
Bill~Yuchen Lin, Xinyue Chen, Jamin Chen, and Xiang Ren. 2019.
\newblock Kagnet: Knowledge-aware graph networks for commonsense reasoning.
\newblock In \emph{Proceedings of the 2019 Conference on Empirical Methods in
  Natural Language Processing and the 9th International Joint Conference on
  Natural Language Processing (EMNLP-IJCNLP)}.

\bibitem[{Liu et~al.(2020)Liu, Cao, Pan, Li, and Chua}]{liu2020exploring}
Zhiyuan Liu, Yixin Cao, Liangming Pan, Juanzi Li, and Tat-Seng Chua. 2020.
\newblock Exploring and evaluating attributes, values, and structure for entity
  alignment.
\newblock In \emph{Proceedings of the 2020 Conference on Empirical Methods in
  Natural Language Processing (EMNLP)}, pages 6355--6364.

\bibitem[{Luo et~al.(2019)Luo, Xu, Zhang, Ren, and Sun}]{pkuseg}
Ruixuan Luo, Jingjing Xu, Yi~Zhang, Xuancheng Ren, and Xu~Sun. 2019.
\newblock Pkuseg: A toolkit for multi-domain chinese word segmentation.
\newblock \emph{CoRR}, abs/1906.11455.

\bibitem[{Mahdisoltani et~al.(2015)Mahdisoltani, Biega
  et~al.}]{mahdisoltani2014yago3}
Farzaneh Mahdisoltani, Joanna Biega, et~al. 2015.
\newblock Yago3: A knowledge base from multilingual {Wikipedias}.
\newblock In \emph{Proceedings of the Conference on Innovative Data Systems
  Research (CIDR)}.

\bibitem[{Manning et~al.(2014)Manning, Surdeanu, Bauer, Finkel, Bethard, and
  McClosky}]{manning2014stanford}
Christopher Manning, Mihai Surdeanu, John Bauer, Jenny Finkel, Steven Bethard,
  and David McClosky. 2014.
\newblock The stanford corenlp natural language processing toolkit.
\newblock In \emph{Proceedings of the Annual Meeting of Associations for
  Computational Linguistics (ACL)}.

\bibitem[{Mikolov et~al.(2013)Mikolov, Chen et~al.}]{mikolov2013efficient}
Tomas Mikolov, Kai Chen, et~al. 2013.
\newblock Efficient estimation of word representations in vector space.
\newblock \emph{International Conference on Learning Representations (ICLR)}.

\bibitem[{Mitchell et~al.(2018)Mitchell, Cohen, Hruschka, Talukdar, Yang,
  Betteridge, Carlson, Dalvi, Gardner, Kisiel et~al.}]{mitchell2018never}
Tom Mitchell, William Cohen, Estevam Hruschka, Partha Talukdar, B~Yang,
  J~Betteridge, A~Carlson, B~Dalvi, M~Gardner, B~Kisiel, et~al. 2018.
\newblock Never-ending learning.
\newblock \emph{Communications of the ACM}.

\bibitem[{Moussallem et~al.(2018)Moussallem, Wauer, and
  Ngomo}]{moussallem2018machine}
Diego Moussallem, Matthias Wauer, and Axel-Cyrille~Ngonga Ngomo. 2018.
\newblock Machine translation using semantic web technologies: A survey.
\newblock \emph{Journal of Web Semantics}, 51:1--19.

\bibitem[{Newman-Griffis et~al.(2018)Newman-Griffis, Lai, and
  Fosler-Lussier}]{newman2018jointly}
Denis Newman-Griffis, Albert~M Lai, and Eric Fosler-Lussier. 2018.
\newblock Jointly embedding entities and text with distant supervision.
\newblock In \emph{Proceedings of the Workshop on Representation Learning for
  NLP}.

\bibitem[{Nickel et~al.(2016)Nickel, Rosasco, Poggio
  et~al.}]{nickel2016holographic}
Maximilian Nickel, Lorenzo Rosasco, Tomaso~A Poggio, et~al. 2016.
\newblock Holographic embeddings of knowledge graphs.
\newblock In \emph{Proceedings of AAAI Conference on Artificial Intelligence
  (AAAI)}.

\bibitem[{Nickel and Kiela(2017)}]{nickel2017poincare}
Maximillian Nickel and Douwe Kiela. 2017.
\newblock Poincar{\'e} embeddings for learning hierarchical representations.
\newblock In \emph{Advances in Neural Information Processing Systems (NIPS)}.

\bibitem[{Pei et~al.(2019{\natexlab{a}})Pei, Yu, Hoehndorf et~al.}]{pei2019deg}
Shichao Pei, Lu~Yu, Robert Hoehndorf, et~al. 2019{\natexlab{a}}.
\newblock semi-supervised entity alignment via knowledge graph embedding with
  awareness of degree difference.
\newblock In \emph{Proceedings of the Web Confererence (WWW)}.

\bibitem[{Pei et~al.(2019{\natexlab{b}})Pei, Yu, and Zhang}]{pei2019transport}
Shichao Pei, Lu~Yu, and Xiangliang Zhang. 2019{\natexlab{b}}.
\newblock Improving cross-lingual entity alignment via optimal transport.
\newblock In \emph{Proceedings of the International Joint Conference on
  Artificial Intelligence (IJCAI)}.

\bibitem[{Pennington et~al.(2014)Pennington, Socher
  et~al.}]{pennington2014glove}
Jeffrey Pennington, Richard Socher, et~al. 2014.
\newblock Glove: Global vectors for word representation.
\newblock In \emph{Proceedings of the Conference on Empirical Methods in
  Natural Language Processing (EMNLP)}.

\bibitem[{Peters et~al.(2018)Peters, Neumann, Iyyer, Gardner, Clark, Lee, and
  Zettlemoyer}]{peters2018deep}
Matthew Peters, Mark Neumann, Mohit Iyyer, Matt Gardner, Christopher Clark,
  Kenton Lee, and Luke Zettlemoyer. 2018.
\newblock Deep contextualized word representations.
\newblock In \emph{Proceedings of the Conference of the North American Chapter
  of the Association for Computational Linguistics: Human Language Technologies
  (NAACL-HLT)}.

\bibitem[{Pires et~al.(2019)Pires, Schlinger, and
  Garrette}]{pires2019multilingual}
Telmo Pires, Eva Schlinger, and Dan Garrette. 2019.
\newblock How multilingual is multilingual bert?
\newblock In \emph{Proceedings of the Annual Meeting of Associations for
  Computational Linguistics (ACL)}.

\bibitem[{Pujara et~al.(2017)Pujara, Augustine, and
  Getoor}]{pujara2017sparsity}
Jay Pujara, Eriq Augustine, and Lise Getoor. 2017.
\newblock Sparsity and noise: Where knowledge graph embeddings fall short.
\newblock In \emph{Proceedings of the Conference on Empirical Methods in
  Natural Language Processing (EMNLP)}.

\bibitem[{Reddi et~al.(2018)Reddi, Kale, and Kumar}]{reddi2018convergence}
Sashank~J Reddi, Satyen Kale, and Sanjiv Kumar. 2018.
\newblock On the convergence of adam and beyond.
\newblock In \emph{International Conference on Learning Representations
  (ICLR)}.

\bibitem[{Rethmeier et~al.(2018)Rethmeier, H{\"u}bner, and
  Hennig}]{rethmeier2018learning}
Nils Rethmeier, Marc H{\"u}bner, and Leonhard Hennig. 2018.
\newblock Learning comment controversy prediction in web discussions using
  incidentally supervised multi-task cnns.
\newblock In \emph{Proceedings of the 9th Workshop on Computational Approaches
  to Subjectivity, Sentiment and Social Media Analysis}.

\bibitem[{Roth(2017)}]{roth2017incidental}
Dan Roth. 2017.
\newblock Incidental supervision: Moving beyond supervised learning.
\newblock In \emph{Proceedings of AAAI Conference on Artificial Intelligence
  (AAAI)}.

\bibitem[{Ruder et~al.(2017)Ruder, Vuli{\'c}, and S{\o}gaard}]{ruder2017survey}
Sebastian Ruder, Ivan Vuli{\'c}, and Anders S{\o}gaard. 2017.
\newblock A survey of cross-lingual word embedding models.
\newblock \emph{Journal of Artificial Intelligence Research}.

\bibitem[{Sch{\"o}nemann(1966)}]{schonemann1966procrustes}
Peter~H Sch{\"o}nemann. 1966.
\newblock A generalized solution of the orthogonal procrustes problem.
\newblock \emph{Psychometrika}, 31(1):1--10.

\bibitem[{Shi et~al.(2019)Shi, Chen, Zhou, and Chang}]{shi2019retrofit}
Weijia Shi, Muhao Chen, Pei Zhou, and Kai-Wei Chang. 2019.
\newblock Retrofitting contextualized word embeddings with paraphrases.
\newblock In \emph{Proceedings of the 2019 Conference on Empirical Methods in
  Natural Language Processing and the 9th International Joint Conference on
  Natural Language Processing (EMNLP-IJCNLP)}.

\bibitem[{Shi and Xiao(2019)}]{shi2019modeling}
Xiaofei Shi and Yanghua Xiao. 2019.
\newblock Modeling multi-mapping relations for precise cross-lingual entity
  alignment.
\newblock In \emph{Proceedings of the 2019 Conference on Empirical Methods in
  Natural Language Processing and the 9th International Joint Conference on
  Natural Language Processing (EMNLP-IJCNLP)}.

\bibitem[{Shvaiko and Euzenat(2011)}]{shvaiko2011ontology}
Pavel Shvaiko and J{\'e}r{\^o}me Euzenat. 2011.
\newblock Ontology matching: state of the art and future challenges.
\newblock \emph{IEEE Transactions on Knowledge and Data Engeering},
  25(1):158--176.

\bibitem[{Sil et~al.(2018)Sil, Ji, Roth, and Cucerzan}]{sil2018multi}
Avirup Sil, Heng Ji, Dan Roth, and Silviu Cucerzan. 2018.
\newblock Multi-lingual entity discovery and linking.
\newblock In \emph{Proceedings of the Annual Meeting of Associations for
  Computational Linguistics (ACL)}.

\bibitem[{Song and Roth(2015)}]{song2015unsupervised}
Yangqiu Song and Dan Roth. 2015.
\newblock Unsupervised sparse vector densification for short text similarity.
\newblock In \emph{Proceedings of the Conference of the North American Chapter
  of the Association for Computational Linguistics: Human Language Technologies
  (NAACL-HLT)}.

\bibitem[{Speer et~al.(2017)Speer, Chin, and Havasi}]{speer2017conceptnet}
Robert Speer, Joshua Chin, and Catherine Havasi. 2017.
\newblock Conceptnet 5.5: An open multilingual graph of general knowledge.
\newblock In \emph{Proceedings of AAAI Conference on Artificial Intelligence
  (AAAI)}.

\bibitem[{Suchanek et~al.(2011)Suchanek, Abiteboul et~al.}]{suchanek2011paris}
Fabian~M Suchanek, Serge Abiteboul, et~al. 2011.
\newblock Paris: Probabilistic alignment of relations, instances, and schema.
\newblock \emph{Proceedings of the VLDB Endowment (PVLDB)}, 5(3).

\bibitem[{Sun et~al.(2019{\natexlab{a}})Sun, Yu, Chen, Yu, Choi, and
  Cardie}]{sun2019dream}
Kai Sun, Dian Yu, Jianshu Chen, Dong Yu, Yejin Choi, and Claire Cardie.
  2019{\natexlab{a}}.
\newblock Dream: A challenge data set and models for dialogue-based reading
  comprehension.
\newblock \emph{Transactions of the Association for Computational Linguistics
  (TACL)}, 7:217--231.

\bibitem[{Sun et~al.(2017)Sun, Hu, and Li}]{sun2017cross}
Zequn Sun, Wei Hu, and Chengkai Li. 2017.
\newblock Cross-lingual entity alignment via joint attribute-preserving
  embedding.
\newblock In \emph{Proceedings of the International Semantic Web Conference
  (ISWC)}.

\bibitem[{Sun et~al.(2018)Sun, Hu, Zhang, and Qu}]{sun2018bootstrapping}
Zequn Sun, Wei Hu, Qingheng Zhang, and Yuzhong Qu. 2018.
\newblock Bootstrapping entity alignment with knowledge graph embedding.
\newblock In \emph{Proceedings of the International Joint Conference on
  Artificial Intelligence (IJCAI)}.

\bibitem[{Sun et~al.(2020{\natexlab{a}})Sun, Wang, Hu, Chen, Dai, Zhang, and
  Qu}]{sun2020alinet}
Zequn Sun, Chengming Wang, Wei Hu, Muhao Chen, Jian Dai, Wei Zhang, and Yuzhong
  Qu. 2020{\natexlab{a}}.
\newblock Knowledge graph alignment network with gated multi-hop neighborhood
  aggregation.
\newblock In \emph{Proceedings of AAAI Conference on Artificial Intelligence
  (AAAI)}.

\bibitem[{Sun et~al.(2019{\natexlab{b}})Sun, Wang, Hu, Chen, and
  Qu}]{sun2019transedge}
Zequn Sun, Jiacheng~Huang Wang, Wei Hu, Muhao Chen, and Yuzhong Qu.
  2019{\natexlab{b}}.
\newblock Transedge: Translating relation-contextualized embeddings for
  knowledge graphs.
\newblock In \emph{Proceedings of the International Semantic Web Conference
  (ISWC)}.

\bibitem[{Sun et~al.(2020{\natexlab{b}})Sun, Zhang, Hu, Wang, Chen, Akrami, and
  Li}]{sun2020benchmark}
Zequn Sun, Qingheng Zhang, Wei Hu, Chengming Wang, Muhao Chen, Farahnaz Akrami,
  and Chengkai Li. 2020{\natexlab{b}}.
\newblock A benchmarking study of embedding-based entity alignment for
  knowledge graphs.
\newblock \emph{Proceedings of the VLDB Endowment (PVLDB)}, 13.

\bibitem[{Sun et~al.(2019{\natexlab{c}})Sun, Deng, Nie, and
  Tang}]{sun2019rotate}
Zhiqing Sun, Zhi-Hong Deng, Jian-Yun Nie, and Jian Tang. 2019{\natexlab{c}}.
\newblock Rotate: Knowledge graph embedding by relational rotation in complex
  space.
\newblock In \emph{International Conference on Learning Representations
  (ICLR)}.

\bibitem[{Tifrea et~al.(2018)Tifrea, Becigneul, and Ganea}]{tifrea2018poincare}
Alexandru Tifrea, Gary Becigneul, and Octavian-Eugen Ganea. 2018.
\newblock Poincare glove: Hyperbolic word embeddings.
\newblock In \emph{International Conference on Learning Representations}.

\bibitem[{Toutanova et~al.(2015)Toutanova, Chen
  et~al.}]{toutanova2015representing}
Kristina Toutanova, Danqi Chen, et~al. 2015.
\newblock Representing text for joint embedding of text and knowledge bases.
\newblock In \emph{Proceedings of the Conference on Empirical Methods in
  Natural Language Processing (EMNLP)}.

\bibitem[{Trsedya et~al.(2019)Trsedya, Qi, and Zhang}]{distiawanTrsedya2019}
Bayu~Distiawan Trsedya, Jianzhong Qi, and Rui Zhang. 2019.
\newblock Entity alignment between knowledge graphs using attribute embeddings.
\newblock In \emph{Proceedings of the AAAI Conference on Artificial
  Intelligence (AAAI)}.

\bibitem[{Upadhyay et~al.(2018)Upadhyay, Gupta, and Roth}]{upadhyay2018joint}
Shyam Upadhyay, Nitish Gupta, and Dan Roth. 2018.
\newblock Joint multilingual supervision for cross-lingual entity linking.
\newblock In \emph{Proceedings of the Conference on Empirical Methods in
  Natural Language Processing (EMNLP)}.

\bibitem[{Veli{\v{c}}kovi{\'c} et~al.(2018)Veli{\v{c}}kovi{\'c}, Cucurull,
  Casanova, Romero, Li{\`o}, and Bengio}]{velivckovic2018gat}
Petar Veli{\v{c}}kovi{\'c}, Guillem Cucurull, Arantxa Casanova, Adriana Romero,
  Pietro Li{\`o}, and Yoshua Bengio. 2018.
\newblock Graph attention networks.
\newblock In \emph{International Conference on Learning Representations
  (ICLR)}.

\bibitem[{Wang et~al.(2014{\natexlab{a}})Wang, Zhang, Feng, and
  Chen}]{wang2014knowledge}
Zhen Wang, Jianwen Zhang, Jianlin Feng, and Zheng Chen. 2014{\natexlab{a}}.
\newblock Knowledge graph embedding by translating on hyperplanes.
\newblock In \emph{Proceedings of AAAI Conference on Artificial Intelligence
  (AAAI)}.

\bibitem[{Wang et~al.(2014{\natexlab{b}})Wang, Zhang et~al.}]{wang2014joint}
Zhen Wang, Jianwen Zhang, et~al. 2014{\natexlab{b}}.
\newblock Knowledge graph and text jointly embedding.
\newblock In \emph{Proceedings of the Conference on Empirical Methods in
  Natural Language Processing (EMNLP)}.

\bibitem[{Wang et~al.(2018)Wang, Lv, Lan, and Zhang}]{wang2018cross}
Zhichun Wang, Qingsong Lv, Xiaohan Lan, and Yu~Zhang. 2018.
\newblock Cross-lingual knowledge graph alignment via graph convolutional
  networks.
\newblock In \emph{Proceedings of the Conference on Empirical Methods in
  Natural Language Processing (EMNLP)}.

\bibitem[{Wijaya et~al.(2013)Wijaya, Talukdar et~al.}]{wijaya2013pidgin}
Derry Wijaya, Partha~Pratim Talukdar, et~al. 2013.
\newblock Pidgin: ontology alignment using web text as interlingua.
\newblock In \emph{Proceedings of the ACM International Conference on
  Information and Knowledge Management (CIKM)}.

\bibitem[{Wu et~al.(2019{\natexlab{a}})Wu, Liu, Feng, Wang, Yan, and
  Zhao}]{wu2019relation}
Yuting Wu, Xiao Liu, Yansong Feng, Zheng Wang, Rui Yan, and Dongyan Zhao.
  2019{\natexlab{a}}.
\newblock Relation-aware entity alignment for heterogeneous knowledge graphs.
\newblock In \emph{Proceedings of the International Joint Conference on
  Artificial Intelligence (IJCAI)}.

\bibitem[{Wu et~al.(2019{\natexlab{b}})Wu, Liu, Feng, Wang, and
  Zhao}]{wu2019jointly}
Yuting Wu, Xiao Liu, Yansong Feng, Zheng Wang, and Dongyan Zhao.
  2019{\natexlab{b}}.
\newblock Jointly learning entity and relation representations for entity
  alignment.
\newblock In \emph{Proceedings of the 2019 Conference on Empirical Methods in
  Natural Language Processing and the 9th International Joint Conference on
  Natural Language Processing (EMNLP-IJCNLP)}.

\bibitem[{Xu et~al.(2019)Xu, Wang, Yu, Feng, Song, Wang, and Yu}]{xu2019cross}
Kun Xu, Liwei Wang, Mo~Yu, Yansong Feng, Yan Song, Zhiguo Wang, and Dong Yu.
  2019.
\newblock Cross-lingual knowledge graph alignment via graph matching neural
  network.
\newblock In \emph{Proceedings of the Annual Meeting of Associations for
  Computational Linguistics (ACL)}.

\bibitem[{Yamada et~al.(2017)Yamada, Shindo, Takeda, and
  Takefuji}]{yamada2017learning}
Ikuya Yamada, Hiroyuki Shindo, Hideaki Takeda, and Yoshiyasu Takefuji. 2017.
\newblock Learning distributed representations of texts and entities from
  knowledge base.
\newblock \emph{Transactions of the Association for Computational Linguistics
  (TACL)}, 5:397--411.

\bibitem[{Yang et~al.(2015)Yang, Yih, He, Gao, and Deng}]{yang2015embedding}
Bishan Yang, Wen-tau Yih, Xiaodong He, Jianfeng Gao, and Li~Deng. 2015.
\newblock Embedding entities and relations for learning and inference in
  knowledge bases.
\newblock \emph{International Conference on Learning Representations (ICLR)}.

\bibitem[{Yang et~al.(2019)Yang, Zou, Shi, Lu, Lin, and Sun}]{yang2019aligning}
Hsiu-Wei Yang, Yanyan Zou, Peng Shi, Wei Lu, Jimmy Lin, and Xu~Sun. 2019.
\newblock Aligning cross-lingual entities with multi-aspect information.
\newblock In \emph{Proceedings of the 2019 Conference on Empirical Methods in
  Natural Language Processing and the 9th International Joint Conference on
  Natural Language Processing (EMNLP-IJCNLP)}.

\bibitem[{Yang et~al.(2020)Yang, Liu, Zhao, Wang, and Xie}]{yangcotsae}
Kai Yang, Shaoqin Liu, Junfeng Zhao, Yasha Wang, and Bing Xie. 2020.
\newblock Cotsae: Co-training of structure and attribute embeddings for entity
  alignment.
\newblock In \emph{Proceedings of AAAI Conference on Artificial Intelligence
  (AAAI)}.

\bibitem[{Yeo et~al.(2018)Yeo, Wang, Cho, Choi, and Hwang}]{yeo2018machine}
Jinyoung Yeo, Geungyu Wang, Hyunsouk Cho, Seungtaek Choi, and Seung-won Hwang.
  2018.
\newblock Machine-translated knowledge transfer for commonsense causal
  reasoning.
\newblock In \emph{Proceedings of AAAI Conference on Artificial Intelligence
  (AAAI)}.

\bibitem[{Zhang et~al.(2019)Zhang, Sun, Hu, Chen, Guo, and Qu}]{zhang2019multi}
Qingheng Zhang, Zequn Sun, Wei Hu, Muhao Chen, Lingbing Guo, and Yuzhong Qu.
  2019.
\newblock Multi-view knowledge graph embedding for entity alignment.
\newblock In \emph{Proceedings of the International Joint Conference on
  Artificial Intelligence (IJCAI)}.

\bibitem[{Zhong et~al.(2015)Zhong, Zhang et~al.}]{zhong2015aligning}
Huaping Zhong, Jianwen Zhang, et~al. 2015.
\newblock Aligning knowledge and text embeddings by entity descriptions.
\newblock In \emph{Proceedings of the Conference on Empirical Methods in
  Natural Language Processing (EMNLP)}.

\bibitem[{Zhu et~al.(2019)Zhu, Zhou, Wu, Tan, and Guo}]{zhu2019neighborhood}
Qiannan Zhu, Xiaofei Zhou, Jia Wu, Jianlong Tan, and Li~Guo. 2019.
\newblock Neighborhood-aware attentional representation for multilingual
  knowledge graphs.
\newblock In \emph{Proceedings of the International Joint Conference on
  Artificial Intelligence (IJCAI)}.

\bibitem[{Zou et~al.(2013)Zou, Socher et~al.}]{zou2013bilingual}
Will~Y Zou, Richard Socher, et~al. 2013.
\newblock Bilingual word embeddings for phrase-based machine translation.
\newblock In \emph{Proceedings of the Conference on Empirical Methods in
  Natural Language Processing (EMNLP)}.

\end{thebibliography}
\bibliographystyle{acl_natbib}

\clearpage
\appendix

\section{Appendices}
\label{sec:appendix}

\subsection{Descriptions of Baseline Methods}\label{sup:base}

We provide descriptions of baseline methods. In accord with \Cref{sec:exp_set}, we also separate the descriptions in three groups.

MTransE~\cite{chen2017multigraph} represents a pioneering method of this topic. It jointly learns a translational embedding model~\cite{bordes2013translating} and an alignment model that captures the correspondence of counterpart entities via  transformations or distances of the embedding representations.
Based on the methodology of MTransE, GCN-Align~\cite{wang2018cross} substitute the translational embedding model with GCN to better capture the entity based on their neighborhood structures.
MECG~\cite{li2019semi} extends the framework of GCN-Align with regularization term based on relational translation, aiming at differentiating between the information neighboring entities that play different roles of relations.
GCN-JE~\cite{wu2019jointly} extends the GCN-Align architecture with embedding calibration also on relations.
MuGCN~\cite{cao2019multi} combines multiple channels of GCNs to better model the heterogeneous neighborhood information of entities in different KGs.
For the same purpose, AliNet~\cite{sun2020alinet} incorporates a gate mechanism in the neighborhood aggregation process of GAT.
Both techniques offer satisfying performance in entity alignment without a transformation between KG-specific embedding spaces.
Different from these neighborhood aggregation techniques, RSN~\cite{guo2019learning} focuses capturing the long-term relational dependency of entities by incorporating a gated recurrent network with highway links, and offers comparable performance to MuGCN.
Besides the above embedding learning techniques, single-graph KG embedding models have also been evaluated for entity alignment in recent studies~\cite{guo2019learning,sun2020alinet}, by simply treating the match of entities as a type of relation in the KG.
According to these studies, while RotatE~\cite{sun2019rotate} outperforms others single-graph embedding models, it is significantly outperformed by most aforementioned entity alignment methods.

Besides different embedding learning techniques, there are approaches to obtain additional supervision signals from profile information of entities that are available in some KBs.
JAPE~\cite{sun2017cross} introduces an auxiliary measure of entity attributes, and use this to strengthen the cross-lingual learning of MTransE.
SEA~\cite{pei2019deg} also obtains similarly auxiliary supervision signals based on centrality measures.
HMAN~\cite{yang2019aligning} is a GCN-based model that incorporates various modalities of entity information, including entity names, attributes, and literal descriptions that are also leveraged in KDCoE~\cite{chen2018co}.

Another line of research focuses on semi-supervised alignment learning to capture the entity alignment based on limited labels.
BootEA~\cite{sun2018bootstrapping}, MMR~\cite{shi2019modeling} and NAEA~\cite{zhu2019neighborhood} similarly conducts an self-learning approach to iteratively propose alignment label on unaligned entities,
The main difference of these three models lies in the embedding learning techniques, given that BootEA is translational, MMR is GCN based, and NAEA is GAT-based.
It is noteworthy that, NAEA also incorporates the mutual nearest neighbor constraint in proposing new alignment labels.
KDCoE adopts an iterative co-training process of MTransE with another self-attentive Siamese encoder of entity descriptions,
and both model components alternately propose alignment labels.
Different from those iterative learning processes, OTEA~\cite{pei2019transport} employs an optimal transport solution that is similar to the Procrustas solution~\cite{schonemann1966procrustes} used in this work.

\subsection{Statistics of the Datasets}\label{sup:stats}

{
\begin{table}[h!]
\setlength\tabcolsep{2pt}
\centering
\scriptsize
\begin{tabular}{c|c|cc}
\bhline
\multicolumn{2}{c|}{Data}&\#Triples&\#Entities\\
\hline
\multirow{2}{*}{DBP15k$_{En-Fr}$}&En&278,590&105,889\\
&Fr&192,191&66,858\\
\hline
\multirow{2}{*}{DBP15k$_{En-Zh}$}&En&237,674&98,125\\
&Zh&153,929&66,469\\
\hline
\multirow{2}{*}{DBP15k$_{En-Ja}$}&En&233,319&95,680\\
&Ja&164,373&65,744\\
\bhline
\end{tabular}
\caption{Statistics of the DBP15k dataset.}\label{tbl:statdbp}
\end{table}
}

{
\begin{table}[h!]
\setlength\tabcolsep{2pt}
\centering
\scriptsize
\begin{tabular}{c|ccc|c|cccc}
\bhline
Data&\#En&\#Fr&\#De&ILL Lang&\#Train&\#Valid&\#Test\\
\hline
\multirow{2}{*}{Triples}&\multirow{2}{*}{569,393}&\multirow{2}{*}{258,337}&\multirow{2}{*}{224,647}&En-Fr&13,050&2,000&39,155\\
&&&&En-De&12,505&2,000&41,018\\
\bhline
\end{tabular}
\caption{Statistics of the WK3l60k dataset.}\label{tbl:statwk}
\end{table}
}

Statistics of the datasets are listed in \Cref{tbl:statdbp} and \Cref{tbl:statwk}.
As described, the partition and use of these datasets are consistent with previous papers~\cite{chen2018co,sun2017cross,sun2018bootstrapping,wang2018cross,yang2019aligning,pei2019transport}. The datasets used in this work are open benchmark datasets. 

\subsection{More Implementation Details}\label{sup:hyper}

\begin{table}[h] 
\small
\setlength{\tabcolsep}{1pt}
\caption{Hyperparameter search spaces.}
\label{tbl:search_space}
\centering
\begin{tabular}{lr}
\bhline
hyper-parameters & search space \\
\hline
Dimension $k$ & \{100, 150, 300, 400\}\\
Bias $b$ in $S^K_L$ & \{1, 2, 4\} \\
Batch sizes & \{128, 256, 512\} \\
Negative sample sizes & \{3, 5, 8\} \\
$k$ in \textsc{Csls} & \{3, 5, 10\}\\
\bhline
\end{tabular}
\end{table}

Our experiments are conducted on a commodity server and use one NVIDIA TITAN RTX 6000 GPU.
On this machine,
each run of the entire training process in the eventual hyperparameter setting takes around 40 minutes.
\Cref{tbl:search_space} takes around list the search spaces where hyper-parameter values are selected according to the validation set provided by the original datasets.
The best configuration is decided based on $\hitsone$.


\end{document}